\documentclass{article}

\usepackage{arxiv}

\usepackage[utf8]{inputenc} 
\usepackage[T1]{fontenc}    
\usepackage{hyperref}       
\usepackage{url}            
\usepackage{booktabs}       
\usepackage{amsfonts}       
\usepackage{nicefrac}       
\usepackage{microtype}      
\usepackage{lipsum}
\usepackage{graphicx}
\graphicspath{ {./images/} }

\title{CycleCap: Improving VLMs Captioning Performance via Self-Supervised Cycle Consistency Fine-Tuning}

\author{
Marios Krestenitis$^{1,2}$
Christos Tzelepis$^{3}$ 
Konstantinos Ioannidis$^{2}$
Stefanos Vrochidis$^{2}$ \\ 
\textbf{Ioannis Kompatsiaris$^{2}$
Georgios Tzimiropoulos$^{1}$
Shaogang Gong$^{1}$
Ioannis Patras$^{1}$}
\\
\\
$^{1}$Queen Mary, University of London \\
$^{2}$Centre for Research and Technology Hellas \\
$^{3}$City St George's, University of London
\\
\texttt{Corresponding author: m.krestenitis@qmul.ac.uk}
}

\usepackage{graphicx}
\usepackage{booktabs}
\usepackage{siunitx}
\usepackage{multirow}
\usepackage{colortbl}
\usepackage{makecell}
\usepackage{adjustbox}
\usepackage[table]{xcolor}
\definecolor{lightgreen}{HTML}{E8F5E9} 
\definecolor{lightgray}{HTML}{F0EDED} 
\usepackage{subcaption}

\usepackage[accsupp]{axessibility}  

\usepackage{hyperref}

\usepackage{orcidlink}

\usepackage{multirow}
\usepackage[table,xcdraw]{xcolor}
\usepackage{amsmath}

\usepackage{soul}

\newcommand{\cyclecaprow}{\rowcolor{lightgreen}}

\let\titleold\title
\renewcommand{\title}[1]{\titleold{#1}\newcommand{\thetitle}{#1}}
\def\maketitlesupplementary
   {
   \newpage
   \begin{center}
    \Large
    Appendix\\[1em]
  \end{center}
   }

\begin{document}
\maketitle


\maketitle

\begin{abstract}
Visual–Language Models (VLMs) have achieved remarkable progress in image captioning, visual question answering, and visual reasoning. Yet they remain prone to vision–language misalignment, often producing overly generic or hallucinated descriptions. Existing approaches address this via instruction tuning-requiring costly, large-scale annotated datasets or via complex test-time frameworks for caption refinement. In this work, we revisit image–text alignment through the lens of cycle consistency: given an image and a caption generated by an image-to-text model, the backward mapping through a text-to-image model should reconstruct an image that closely matches the original. In our setup, a VLM serves as the image-to-text component, while a pre-trained text-to-image model closes the loop by reconstructing the image from the generated caption. Building on this, we introduce CycleCap, a fine-tuning scheme to improve image captioning using Group Relative Policy Optimization (GRPO) with a reward based on the similarity between the original and reconstructed images, computed on-the-fly. Unlike previous work that uses cycle consistency loss for preference dataset construction, our method leverages cycle consistency directly as a self-supervised training signal. This enables the use of raw images alone, eliminating the need for curated image–text datasets, while steering the VLM to produce more accurate and grounded text descriptions. Applied to four VLMs ranging from 1B to 7B parameters, CycleCap yields consistent improvements across captioning and hallucination benchmarks, surpassing state-of-the-art methods that rely on supervised cycle consistency training. 

\end{abstract}

\section{Introduction}
\label{sec:intro}

\begin{figure*}[t]
    \centering
    \includegraphics[width=\textwidth]{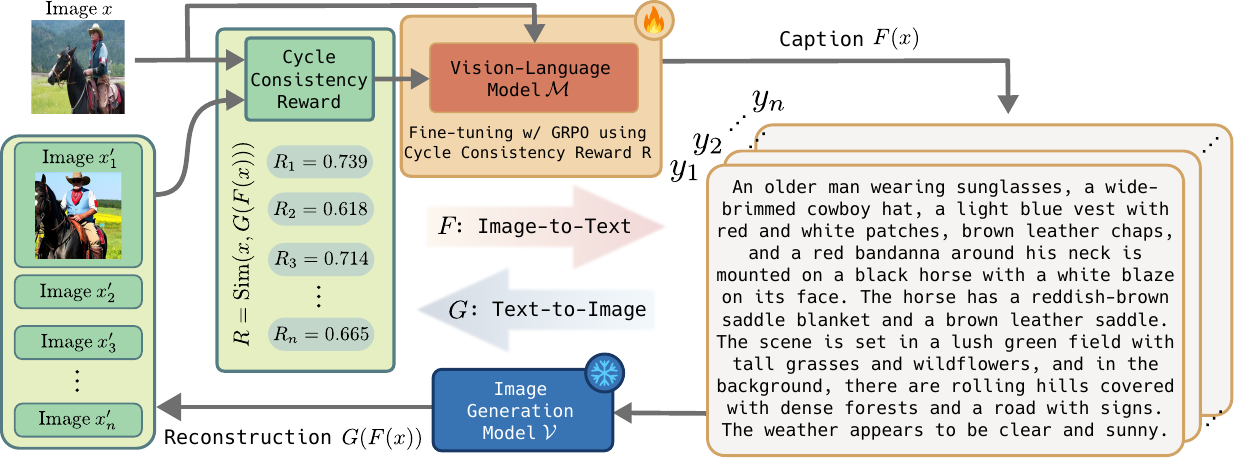}
    \caption{Overview of the CycleCap framework. The Visual-Language Model generates multiple captions 
    $\{y_i\}_{i=1}^{n}$ for an image $x$. Each caption is used by a frozen Image Generation Model to reconstruct an image $x'_i = G(y_i)$, whilst the similarity between $x$ and $G(y_i)$ is measured to obtain the 
    cycle consistency reward $R_i$, $i=1,\ldots,n$. These rewards guide fine-tuning of the Visual--Language Model via GRPO to encourage captions that better reflect visual content in a self-supervised manner.}
    \label{fig::cyclecap_framework}
\end{figure*}

Visual-Language Models (VLMs) are rapidly advancing as effective methods for understanding visual and textual information. Their ability to process both modalities has them demonstrating remarkable performance in several tasks, such as image captioning, visual question answering and visual reasoning~\cite{li2023blip,alayrac2022flamingo,dai2023instructblip, wang2024qwen2,li2024llava}. Despite the impressive results, VLMs still face significant challenges in effectively aligning vision and language~\cite{yarom2023you, bai2024hallucination, liu2024survey}. Images convey concrete structural and perceptual details that are often difficult to fully express in text, while text captures abstract semantics that may lack direct visual counterparts. Rethinking text-to-image and image-to-text generations as translation tasks, the mapping between images and text is inherently many-to-many rather than one-to-one due to this semantic ambiguity. Such misalignment between modalities can lead to overly generic descriptions or even hallucinated content that fails to accurately ground visual information.

Several approaches have been proposed to mitigate this issue, including instruction tuning strategies and reinforcement learning with human or automated feedback~\cite{yuan2024self,li2025generation,yu2024rlhf, ziegler2019fine}. However, these methods often require large-scale paired datasets, typically obtained through costly human annotation~\cite{bai2022training, yu2024rlhf}. Another line of work focuses on refining pre-acquired captions at inference time. To this end, multi-stage frameworks have been developed that integrate VLMs, Large Language Models (LLMs), and supplementary detectors to enhance text descriptions and reduce hallucinations~\cite{peng2025patch, dong2024benchmarking, pi2024image}. Despite bypassing additional training, these approaches rely on complex and expensive pipelines, while the extensive computation required at inference time limits their scalability.

In this paper, we revisit image-text alignment through the lens of cycle consistency~\cite{zhu2017unpaired}. As illustrated in Fig.~\ref{fig::cyclecap_framework}, given an image $x\in\mathcal{X}$ and the image-to-text and text-to-image mappings $F\colon\mathcal{X}\to\mathcal{Y}$ and $G\colon\mathcal{Y}\to\mathcal{X}$, respectively, the translation between the two modalities is considered accurate when $G(F(x))\approx x$, where the accuracy is measured by the similarity between $x$ and $G(F(x))$. Our key idea is to use cycle consistency directly as a self-supervised training signal to improve image-text alignment. In this light, we introduce CycleCap, a cycle-consistent fine-tuning framework for VLMs that adapts Group Relative Policy Optimization (GRPO)~\cite{shao2024deepseekmath} and incorporates a reward function that encourages greater similarity between $x$ and $G(F(x))$. This steers the model to generate more accurate and detailed descriptions, improving image-text alignment and visual grounding.

Our work differs from previous approaches that have explored the use of cycle consistency primarily as an evaluation or post-process refinement tool for text descriptions~\cite{cui2025evaluating, huang2025image2text2image, chancycle, diesendruck2024learning}. These methods have shown that higher cycle consistency correlates with more accurate image-to-text generations, but they typically treat it as an outcome of the model rather than a training objective. It also differs from studies that have leveraged cycle consistency loss to construct image-text preference datasets~\cite{wang2025rico, bahng2025cycle}, where ensembles of image-to-text and text-to-image models were used to rank image-text pairs for supervised fine-tuning of VLMs with Direct Preference Optimization (DPO)~\cite{rafailov2023direct}. 

In contrast, our approach employs cycle consistency directly as a self-supervised learning signal during training. The proposed GRPO reinforcement learning scheme enforces cycle consistency on-the-fly, allowing the model to learn from image input alone and improve image-text alignment without relying on expensive preference datasets or post-processing refinements. We apply our method to fine-tune four novel VLMs of varying sizes, ranging from 1B to 7B parameters, and demonstrate consistent performance improvements across captioning and hallucination benchmarks. Our method surpasses current state-of-the-art approaches that leverage cycle consistency loss for supervised training. Our main contributions are summarized as follows:
\begin{itemize}
    \item We introduce CycleCap, a novel cycle consistent fine-tuning framework for VLMs that uses the power of GRPO learning to acquire dense and accurate image-to-text outputs. Our approach leverages the idea of cycle consistency, allowing the use of only image data instead of requiring curated and expensive image-text tuning datasets.
    \item We deploy our method for four VLMs of varying size, ranging from 1B to 7B parameters, and demonstrate consistent performance improvements across captioning and hallucination benchmarks.
    \item We evaluate our method against state-of-the-art approaches that rely on cycle consistency to construct preference datasets for fine-tuning, and show that it surpasses these methods while operating in a fully self-supervised manner, without requiring costly and time-consuming dataset creation.
    \item We conduct extensive ablations showing that CycleCap is robust to different image-similarity metrics and text-to-image backbones, and that stronger reconstruction models or perceptual metrics yield higher gains, demonstrating the flexibility and scalability of the framework.
\end{itemize}

\section{Related Work}
\label{sec:rel_work}

\subsection{Visual-Language Models}
Visual–Language models (VLMs) integrate visual and textual understanding within a unified architecture, making them the foundation of multimodal understanding and reasoning. Early approaches, such as CLIP~\cite{radford2021learning} and ALIGN~\cite{jia2021scaling}, established large-scale image–text alignment through contrastive learning, while models like BLIP-2~\cite{li2023blip} bridged pretrained visual encoders and language models using lightweight adapters. Since then, research has shifted toward building instruction-tuned and conversational foundation models that align multimodal behavior with natural language instructions. In this light, several multimodal architectures have been introduced, such as MiniGPT-4~\cite{zhu2023minigpt}, LLaVA~\cite{liu2023visual}, Qwen-VL~\cite{bai2023qwen}, Gemini~\cite{team2023gemini}, and GPT4V~\cite{OpenAI_2023_GPT4V_systemcard}, that extend pretrained LLMs (e.g., LLaMa~\cite{touvron2023llama} or Qwen~\cite{bai2023qwen}) with visual encoders and multimodal alignment training, enabling more grounded and context-aware responses. Parallel work explores post-training optimization methods—including reinforcement learning with human or automated feedback~\cite{christiano2017deep,lee2023rlaif,bai2022constitutional}, or more recent DPO~\cite{rafailov2023direct} and GRPO~\cite{shao2024deepseekmath}, to improve models' reasoning, alignment with human preferences, and instruction-following capabilities through robust post-training methods.

\subsection{Detailed Image Captioning}
Despite their scale and versatility, current VLMs still face challenges in achieving fine-grained visual grounding and rich descriptive detail. Producing detailed image descriptions is a challenging task that requires fine spatial reasoning, attribute recognition, and the ability to avoid hallucinated or generic content. A key factor behind this challenge lies in the large-scale image–text datasets~\cite{schuhmann2022laion, agrawal2019nocaps} used for pretraining, which often contain short or noisy text descriptions that limit the models' ability to learn rich and fine-grained visual–textual correspondences. To overcome this issue, researchers have explored enriching these datasets by re-captioning them with more detailed descriptions, as in ShareGPT4V~\cite{chen2024sharegpt4v} and IIW~\cite{garg2024imageinwords}, or have focused on inference-time methods~\cite{peng2025patch, pi2024image} to improve caption detail at test time. Evaluating detailed captions is also challenging, since traditional reference-based metrics such as BLEU~\cite{papineni2002bleu} and CIDEr~\cite{vedantam2015cider} are sensitive to writing style and struggle with long, descriptive sentences. To address this, DetailCaps4870~\cite{lu2025benchmarking}, CompreCap~\cite{lu2025benchmarking}, CapsBench~\cite{liu2024playground}, and CAPability~\cite{liu2025capability} propose more comprehensive evaluation protocols, often combining LLM-based judging with multilevel alignment metrics. Although these advances improve both caption quality and its assessment, generating accurate and richly detailed captions without additional supervision remains an open challenge.

\subsection{Cycle consistency}
\label{sec::related_cycle_consistency}

Cycle consistency, originally introduced by CycleGAN~\cite{zhu2017unpaired} for unpaired image-to-image translation, has been applied across diverse tasks~\cite{godard2017unsupervised, wang2019unsupervised, hoffman2018cycada,yi2017dualgan}. Recently, it has been extended to VLMs and multimodal systems. Recent works~\cite{huang2025image2text2image,cui2025evaluating,chancycle} have proposed using cycle consistency as a metric to evaluate the captioning performance of image-to-text models, based on the idea that a faithful text description should enable accurate reconstruction of the original image. Beyond evaluation, Li et al.~\cite{li2023leveraging} leveraged cycle consistency for vision–language learning by designing a specialized architecture trained with token-level reconstruction loss to exploit unpaired image-text data. Although the approach was introduced to reduce reliance on large paired datasets, it still required approximately 3M of paired image–text examples for initialization. Moreover, the cycle consistency objective was tightly integrated into the model architecture, making the method dependent on the specialized design rather than serving as a training scheme. On the other hand,  CyclePrompt~\cite{diesendruck2024learning} employed a cyclical framework to refine captions offline, generating images from the captions of original images with DALL-E 3~\cite{betker2023improving} and using GPT-4V~\cite{OpenAI_2023_GPT4V_systemcard} to improve those captions based on differences between the original and the reconstructed images. 

Building on this idea, Wang et al.~\cite{wang2025rico} introduced the RICO, an iterative caption refinement process that uses FLUX.1-dev~\cite{labs2025flux1kontextflowmatching} to reconstruct images from captions and GPT-4o~\cite{openai_gpt4o_2024} to refine captions based on discrepancies between the original and reconstructed images. This iterative process is used to construct a preference dataset of caption pairs (initial versus refined) to train the captioning model RICO-Flash. Similarly, Bahng et al.~\cite{bahng2025cycle} proposed the CyclePref framework, a cycle-based preference construction pipeline. They used an ensemble of 11 image-to-text models (0.5B–40B parameters) to generate caption variations and Stable Diffusion 3 (SD3)~\cite{esser2024scaling} for image reconstruction. Generated captions were ranked by comparing the image reconstructions to the original inputs, forming the CyclePrefDB-I2T preference dataset to improve captioning performance.

Despite these advancements, previous works either treat cycle consistency as an evaluation characteristic or use it as a ranking criterion to build costly preference datasets. In contrast, our approach, CycleCap, leverages cycle consistency directly as a self-supervision signal to improve VLM performance and enhance captioning capabilities without requiring annotated image-text pairs or relying on multiple large-scale components or external APIs such as GPT-4.

\section{CycleCap Framework}
\label{sec:cyclecap}

In this section, we present the CycleCap framework for fine-tuning VLMs to improve image captioning quality. As shown in Fig.~\ref{fig::cyclecap_framework}, CycleCap is a simple yet effective approach built on two complementary components: 
\begin{itemize}
    \item An image-to-text VLM model $\mathcal{M}$ denoted as $F\colon\mathcal{X}\to\mathcal{Y}$ that maps images $x\in\mathcal{X}$ to captions $y\in\mathcal{Y}$.
    \item An image generation model $\mathcal{V}$ that performs the reverse mapping $G\colon\mathcal{Y}\to\mathcal{X}$.
\end{itemize}
Our objective is to fine-tune $\mathcal{M}$ to generate textual descriptions that are better grounded to the visual content of an image.

\paragraph{\textbf{Cycle Consistency Reward}} Given an image $x\in\mathcal{X}$, the VLM model $\mathcal{M}$ produces a textual description $y=F(x)\in\mathcal{Y}$. To evaluate how well the text reflects the image content, we measure how accurately it can reconstruct the original image through $G(y)\in\mathcal{X}$. To this end, we define the cycle consistency reward as
\begin{equation}\label{eq:cycle_reward}
R = \text{Sim}(x, G(F(x))),
\end{equation}
where $\text{Sim}(\cdot, \cdot)$ denotes a similarity metric between the original and reconstructed image. In our experiments, we measure this similarity with DreamSim~\cite{fu2023dreamsim}, which assesses the perceptual and semantic correspondence between images. This formulation allows for estimating the quality of the generated text descriptions without requiring reference captions. A caption that enables accurate reconstruction of the input image implicitly captures visual semantics and structure, and thus (\ref{eq:cycle_reward}) serves as a self-supervised measure of description quality that we later use to guide model fine-tuning.

\paragraph{\textbf{Fine-tuning with GRPO}} We adopt the cycle consistency reward (\ref{eq:cycle_reward}) to fine-tune $\mathcal{M}$ with GRPO. The objective is to maximize $R$, encouraging the model to generate captions that better reflect the visual content. During training, only the parameters of $\mathcal{M}$ are updated, while the text-to-image model $\mathcal{V}$ remains frozen.

As illustrated in Figure \ref{fig::cyclecap_framework}, for each image $x\in\mathcal{X}$, the VLM generates a group of $n$ candidate captions $y_i\in\mathcal{Y}$, $i=1,\ldots,n$. Each caption is fed to the image generator $\mathcal{V}$ to generate the reconstructed image $x_i^\prime=G(y_i)\in\mathcal{X}$. Each caption $y_i$ is assigned with a similarity score $R_i = \text{Sim}(x, G(y_i))$. Then, the relative advantage of each caption within the group is calculated as
\begin{equation}
    A_i = \frac{R_i - \bar{R}}{s_R},
\end{equation}\label{eq::advantage}
where $\bar{R}$ and $s_R$ are the mean and standard deviation of the rewards $\{R_1,\ldots,R_n\}$ within the group, respectively. The GRPO loss is then defined as
\begin{equation}\label{eq::loss_grpo}
\mathcal{L}_{\text{GRPO}} =
-\,\mathbb{E}\left[
\frac{1}{n}\sum_{i=1}^{n}
\min\left(
\rho_i(\theta) A_i,\;
\mathrm{clip}(\rho_i(\theta),\,1-\varepsilon,\,1+\varepsilon)\,A_i
\right)
\right] 
+ \beta\,
D_{\mathrm{KL}}\left(
\pi_\theta \,\|\, \pi_{\mathrm{ref}}
\right),
\end{equation}
where $\rho_i = \frac{\pi_\theta(y_i \mid x)}{\pi_{\theta_{\text{old}}}(y_i \mid x)}$ is the likelihood ratio between the updated policy $\pi_\theta$ and the previous policy $\pi_{\theta_{\text{old}}}$, $\varepsilon$ is the clipping threshold and $\beta$ the weight of the KL regularization term used to constrain the updated policy $\pi_\theta$ to stay close to a frozen reference policy $\pi_{\text{ref}}$. This encourages the model to increase the likelihood of captions that yield higher reconstruction rewards, leading to more accurate text descriptions, while preventing divergence from the original model behavior.

\section{Experiments}
\label{sec:experiments}

In this section, we conduct extensive comparisons with widely used VLMs and benchmarks to evaluate the performance gains of our approach.

\subsection{Implementation Details}
\label{sec:implementation_details}

We deploy CycleCap to fine-tune a set of vision–language models of varying sizes, ranging from 1B to 7B parameters -- specifically, InternVL3-1B~\cite{zhu2025internvl3}, Qwen2-VL-2B~\cite{wang2024qwen2}, Qwen2.5-VL-3B~\cite{bai2025qwen2_5}, and Qwen2-VL-7B~\cite{wang2024qwen2}. Training is performed on the COCO 2014 train split~\cite{lin2014microsoft}, which contains approximately 83,000 everyday scene images. All models are fine-tuned for one epoch with a learning rate of $10^{-5}$, a global batch size of 64, and the number of GRPO rollouts (caption generations per image) is set to 8. For efficiency, we apply LoRA adaptation~\cite{hu2022lora} with rank 64. During training, each model is prompted to produce a detailed description of the image; the exact prompt template is provided in Fig.~\ref{fig::cyclecap_prompt} in the Appendix. For the backward text-to-image mapping $G\colon\mathcal{Y}\to\mathcal{X}$, we employ Stable Diffusion 3 (SD3)~\cite{esser2024scaling} similarly to CyclePref~\cite{bahng2025cycle} with its default parameters. Moreover, for a fair comparison with RICO~\cite{wang2025rico}, we additionally employ FLUX.1-dev~\cite{labs2025flux1kontextflowmatching} text-to-image generator. To mitigate variance in the reward signal introduced by stochastic image generation, we fix the random seed per image sample. All models were trained using 2 A100 GPUs.

\subsection{Evaluation Benchmarks}
\label{benchmarks}
We study the effect of CycleCap fine-tuning on model performance, mainly on the captioning task. To this end, we employ a set of widely used multimodal benchmarks that assess detailed image captions in terms of completeness and correctness. Specifically, we use CompreCap~\cite{lu2025benchmarking}, which contains 560 annotated images and measures caption quality from a structured scene-graph perspective by evaluating object-level coverage as well as the accuracy of object attributes and inter-object relations. Moreover, we use CAPability~\cite{liu2025capability}, a comprehensive benchmark that evaluates captions across multiple aspects of an image (i.e., object category, number, and color, spatial relations, scene, camera angle, OCR, style and character identification), each containing approx. 1,000 human-annotated images, and checks whether captions correctly and thoroughly describe those elements compared to ground-truth annotations using a GPT-based evaluator. We also use CapsBench~\cite{liu2024playground}, which contains 200 images and tests whether generated captions provide sufficient visual grounding by prompting an LLM to answer ``yes/no'' questions about the image, based solely on the caption. Furthermore, we examine the model performance in terms of hallucinations as a critical aspect that affects caption accuracy. To this end, we use MMHal~\cite{sun2024aligning}, which contains 96 images paired with challenging image-related questions and scores the responses using an LLM on a $0$–$6$ scale, where $0$ denotes fully hallucinated and $6$ indicates highly accurate and well-reasoned answers. 

For the CAPability benchmark, we use the official prompt to generate image captions. For CompreCap and CapsBench, we extract captions from the evaluated models using the official CompreCap prompt. In the cases of GPT-based evaluators, we employ GPT-4o-mini~\cite{openai2024gpt4omini}. For the smaller benchmarks, CapsBench and MMHal, the evaluation is repeated three times to ensure consistency.

\subsection{Comparison with Baseline Models}
\label{comparison_with_vanilla}

Table~\ref{table::comparison_vanilla_captioning} summarizes the improvements in captioning performance achieved by the proposed CycleCap fine-tuning compared to the corresponding baseline models. In these experiments, we use the standard SD3~\cite{esser2024scaling} as an image generator. As shown, CycleCap consistently improves caption quality across all benchmarks and model sizes. On CompreCap~\cite{lu2025benchmarking}, it yields 2–3$\%$ improvements on the Unified Score. Improvements on CAPability~\cite{liu2025capability}, a more challenging benchmark that holistically examines the model's captioning abilities across multiple evaluation dimensions, further confirm that CycleCap leads to captions that are more informative, semantically complete, and thoroughly aligned with diverse aspects of the visual content. Similarly, CycleCap leads to improvements of over $2\%$ in most cases, when evaluated on CapsBench~\cite{liu2024playground}. Accordingly, gains in MMHal~\cite{sun2024aligning} indicate that the generated descriptions are more informative while containing less hallucinations. Notably, these gains are observed even for the larger model Qwen2-VL-7B, suggesting that CycleCap complements existing large-scale multimodal pretraining rather than merely compensating for smaller model architectures. These results confirm that using the cyclic consistency reward benefits the model’s ability to generate detailed and faithful image descriptions.

\begin{table}[t]
\centering
\caption{Comparison of the proposed CycleCap with different baseline models (InternVL3-1B~\cite{zhu2025internvl3}, Qwen2-VL-2B~\cite{wang2024qwen2}, Qwen2.5-VL-3B~\cite{bai2025qwen2_5}, Qwen2-VL-7B~\cite{wang2024qwen2}) on captioning (CompreCap~\cite{lu2025benchmarking}, CAPability~\cite{liu2025capability}, CapsBench~\cite{liu2024playground}) and hallucination (MMHal~\cite{sun2024aligning}) benchmarks.}
\label{table::comparison_vanilla_captioning}
\setlength{\tabcolsep}{5pt}
\renewcommand{\arraystretch}{1.15}
\footnotesize
\begin{adjustbox}{max width=\textwidth}
\begin{tabular}{
l
S[table-format=2.2]
S[table-format=1.2]
S[table-format=1.2]
S[table-format=2.2]
S[table-format=2.2]
S[table-format=2.2]
S[table-format=1.2]
}
\toprule
& \multicolumn{4}{c}{CompreCap~\cite{lu2025benchmarking}} 
& \multicolumn{1}{c}{\makecell{CAPability~\cite{liu2025capability}\\$[0,100]$}}
& \multicolumn{1}{c}{\makecell{CapsBench~\cite{liu2024playground}\\$[0,100]$}}
& \multicolumn{1}{c}{\makecell{MMHal~\cite{sun2024aligning}\\$[0,6]$}} \\
\cmidrule(lr){2-5}
Model
& {\makecell{Object Coverage\\$[0,100]$}}
& {\makecell{Attribute Score\\$[0,5]$}}
& {\makecell{Relation Score\\$[0,5]$}}
& {\makecell{Unified Score\\$[0,100]$}}
& {} & {} & {} \\
\midrule

\multicolumn{8}{c}{\textit{InternVL3-1B}~\cite{zhu2025internvl3}} \\ \hline

Baseline                & 73.35 & 2.75 & 2.83 & 60.24 & 69.37 & 71.93 & 3.29 \\
\cyclecaprow CycleCap (Ours) & \bfseries 77.35 & \bfseries 2.89 & \bfseries 2.87 & \bfseries 62.49 & \bfseries 70.89 & \bfseries 73.37 & \bfseries 3.36 \\
\addlinespace[3pt]
\hline
\multicolumn{8}{c}{\textit{Qwen2-VL-2B}~\cite{wang2024qwen2}} \\ \hline
Baseline                & 73.10 & 2.72 & 2.76 & 59.35 & 69.20 & 69.70 & 3.63 \\
\cyclecaprow CycleCap (Ours) & \bfseries 76.94 & \bfseries 2.85 & \bfseries 2.86 & \bfseries 62.09 & \bfseries 70.96 & \bfseries 72.11 & \bfseries 3.71 \\
\addlinespace[3pt]
\hline
\multicolumn{8}{c}{\textit{Qwen2.5-VL-3B}~\cite{bai2025qwen2_5}} \\ \hline
Baseline                & 72.30 & 2.72 & 2.76 & 59.21 & 68.70 & 69.52 & 3.78 \\
\cyclecaprow CycleCap (Ours) & \bfseries 76.92 & \bfseries 2.87 & \bfseries 2.88 & \bfseries 62.42 & \bfseries 71.45 & \bfseries 73.56 & \bfseries 4.09 \\
\addlinespace[3pt]
\hline
\multicolumn{8}{c}{\textit{Qwen2-VL-7B}~\cite{wang2024qwen2}} \\ \hline
Baseline                & 77.86 & 2.84 & 2.80 & 61.73 & 70.47 & 74.17 & 3.85 \\
\cyclecaprow CycleCap (Ours) & \bfseries 79.32 & \bfseries 2.91 & \bfseries 2.86 & \bfseries 63.06 & \bfseries 72.95 & \bfseries 76.38 & \bfseries 4.02 \\

\bottomrule
\end{tabular}
\end{adjustbox}
\end{table}

\begin{figure}[t]
  \centering
  \begin{subfigure}[b]{0.49\textwidth}
    \centering
    \includegraphics[width=\textwidth]{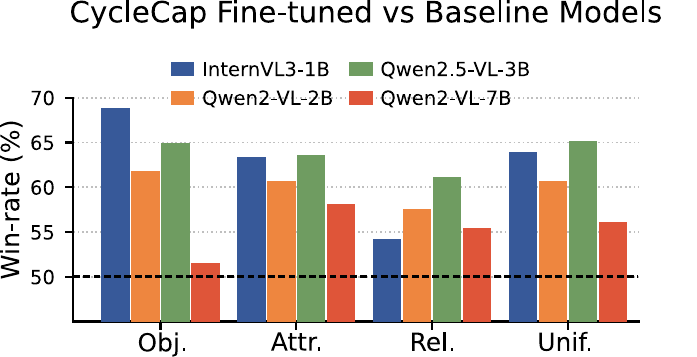}
    \caption{}
    \label{fig:winrate_cyclecap_vs_baseline}
  \end{subfigure}
  \begin{subfigure}[b]{0.45\textwidth}
    \centering
    \includegraphics[width=\textwidth]{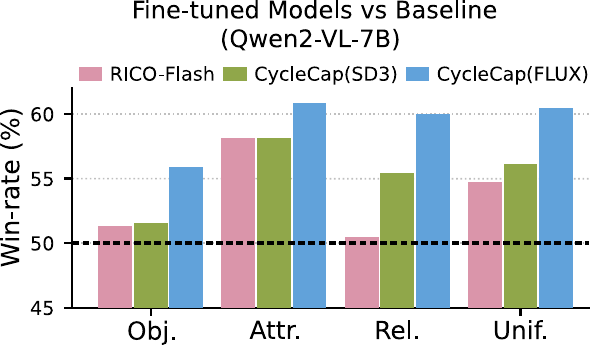}
    \caption{}
    \label{fig:winrate_ft_models_vs_baseline}
  \end{subfigure}
  \caption{Win-rates (\%) of (a) CycleCap fine-tuned models versus the corresponding baseline and (b) Qwen2-VL-7B~\cite{wang2024qwen2} fine-tuned with different methods versus the baseline for CompreCap~\cite{lu2025benchmarking} benchmark.}
  \label{fig:winrate_plots}
\end{figure}

Furthermore, in Fig.~\ref{fig:winrate_cyclecap_vs_baseline}, we show per-image win-rates between fine-tuned and baseline models on CompreCap, indicating how often CycleCap outperforms the baseline under the same metric. We show that CycleCap exceeds the baseline in $>50\%$ of cases across all metrics, indicating consistent improvements at the benchmark level rather than gains concentrated on a small subset of examples. While the relative win-rates are particularly pronounced for smaller models (1B–3B), especially in object coverage and attribute scores, improvements are consistent across all model scales and benchmarks (see Tab.~\ref{table::comparison_vanilla_captioning}).

\subsection{Comparison with State-of-the-Art (SOTA)}
\label{sec::comparison_soa}

In this section, we compare CycleCap with SOTA methods that leverage cycle consistency to improve captioning performance. More specifically, we compare with RICO-Flash~\cite{wang2025rico}, a version of Qwen2-VL-7B~\cite{wang2024qwen2} fine-tuned on a 10K preference dataset derived through the RICO framework, an iterative caption refinement process that uses GPT-4o~\cite{hurst2024gpt} feedback by comparing the original with one reconstructed from its caption using FLUX.1-dev~\cite{labs2025flux1kontextflowmatching}. Moreover, we compare with CyclePref~\cite{bahng2025cycle}, which constructs a larger 398K image-text preference dataset (called CyclePrefDB-I2T) using an ensemble of 11 image-to-text models, ranging from 0.5 to 40B parameters. Caption preference is ranked based on the similarity between the input image and an image reconstructed from its caption using Stable Diffusion 3 (SD3)~\cite{esser2024scaling}. Then, CyclePrefDB-I2T is used to fine-tune Qwen-VL-7B-Chat~\cite{bai2023qwen} for the captioning task with DPO. In our experiments, we follow the setup of \cite{bahng2025cycle} and fine-tune Qwen2-VL-7B with DPO on the CyclePrefDB-I2T dataset.

In Tab.~\ref{table::soa_comparison} we summarize the comparisons with SOTA. The proposed CycleCap consistently achieves higher performance across all benchmarks compared to supervised fine-tuning methods. Notably, CycleCap also improves performance on the CAPability~\cite{liu2025capability} benchmark, a more challenging evaluation where prior methods show limited or even negative gains over the baseline. Overall, CycleCap with SD3 improves the SOTA by a margin comparable to the gap previously observed between the baseline and prior SOTA approaches~\cite{wang2025rico, bahng2025cycle}. It surpasses RICO-Flash~\cite{wang2025rico} and CyclePref~\cite{bahng2025cycle} in caption completeness and factual grounding, demonstrating that direct cyclic self-supervision provides a more effective training signal than preference-based refinement approaches. On top of that, using FLUX.1-dev for text-to-image generation, similarly to RICO-Flash~\cite{wang2025rico}, leads to additional performance gains. This trend is also reflected in Fig.~\ref{fig:winrate_ft_models_vs_baseline}, which reports per-image win-rates against the baseline Qwen2-VL-7B for RICO-Flash and CycleCap (with SD3 and FLUX.1-dev respectively) under the CompreCap metrics. CycleCap variants achieve consistently higher win-rates, indicating more consistent improvements relative to the baseline compared to RICO-Flash.


Moreover, in Tab.~\ref{tab:capability_comparison} we compare CycleCap to SOTA across each evaluation dimension of the CAPability~\cite{liu2025capability}  benchmark, providing a more fine-grained assessment. CycleCap achieves the best overall performance, with the FLUX.1-dev variant obtaining the highest average score. While prior cycle-based approaches show limited or inconsistent gains over the baseline, CycleCap consistently improves performance across all dimensions. Notably, our method demonstrates substantial improvements particularly in object category and number recognition, scene understanding, style description, and character identification -- dimensions that require precise semantic understanding and strong visual grounding. The consistent gains across SD3 and FLUX.1-dev variants indicate that CycleCap enhances caption quality by generating richer descriptions that capture diverse visual attributes and relational aspects of the image.


Overall, Tables \ref{table::soa_comparison} and \ref{tab:capability_comparison} highlight the effectiveness of CycleCap. Notably, our method requires only an image generator, whereas competing SOTA methods require additional substantially heavier external components apart from the image generator itself. Specifically, CyclePref~\cite{bahng2025cycle} is built upon an ensemble of large and costly image-to-text models (up to 40B parameters), while RICO-Flash~\cite{wang2025rico} depends on iterative refinement using the costly GPT-4o. In both cases, computational cost scales with performance, as improvements to the preference-based supervision are inherently tied either to increasing the number and scale of ensemble models or to performing additional caption refinement iterations through GPT-4o API calls. By contrast, CycleCap replaces static preference-based supervision with a continuous, image-grounded feedback that allows the model to directly optimize for image-text alignment using only raw images rather than imitating preference judgments and without relying on external API calls or multiple heavyweight components. Using CycleCap's GRPO optimization scheme encourages progressive improvements through exploration of multiple caption hypotheses, allowing the model to refine its outputs beyond the limitations of static preference pairs and without scaling external supervision complexity.

\begin{table*}[t]
\centering
\caption{Comparison of CycleCap with state-of-the-art cycle-consistency-based methods on CompreCap~\cite{lu2025benchmarking}, CAPability~\cite{liu2025capability}, CapsBench~\cite{liu2024playground}, and MMHal~\cite{sun2024aligning} benchmarks, using Qwen2-VL-7B~\cite{wang2024qwen2} as baseline. For CompreCap we report here only Unified Score, the analytic table is included in the Appendix.}
\label{table::soa_comparison}
\setlength{\tabcolsep}{8pt}
\resizebox{\textwidth}{!}{
\begin{tabular}{l c c c c c}
\toprule
Method & T2I Model & 
CompreCap & 
CAPability &
CapsBench & 
MMHal \\
\midrule
\rowcolor{lightgray} Baseline (Qwen2-VL-7B) & --- & 61.73 & 70.47 & 74.17 & 3.85 \\
CyclePref~\cite{bahng2025cycle} & SD3 & 62.03 & 70.59 & 74.27 & 3.95 \\
RICO-Flash~\cite{wang2025rico} & FLUX.1-dev & 62.93 & 68.83 & 75.30 & 3.92 \\
\midrule
\cyclecaprow CycleCap (Ours) & SD3 & 63.06 & 72.95 & 76.38 & 4.02 \\
\cyclecaprow CycleCap (Ours) & FLUX.1-dev & \textbf{63.64} & \textbf{73.73} & \textbf{77.25} & \textbf{4.02} \\
\bottomrule
\end{tabular}
}
\end{table*}

\begin{table}[t]
\centering
\caption{Comparison of CycleCap with state-of-the-art methods on the CAPability~\cite{liu2025capability} benchmark, all using Qwen2-VL-7B~\cite{wang2024qwen2} as baseline. Bold and underlined values indicate the best and second best results, respectively.}
\resizebox{\textwidth}{!}{
\footnotesize
\setlength{\tabcolsep}{4pt}
\sisetup{table-number-alignment=center}
\begin{tabular}{@{}l
>{\columncolor{lightgray}}S[table-format=2.2]
S[table-format=2.2]
S[table-format=2.2]
|
>{\columncolor{lightgreen}}S[table-format=2.2]
>{\columncolor{lightgreen}}S[table-format=2.2]}
\toprule
Metric 
& {\shortstack{Baseline\\\scriptsize (Qwen2-VL-7B~\cite{wang2024qwen2})}} 
& {\shortstack{CyclePref~\cite{bahng2025cycle}\\\scriptsize (SD3)}}
& {\shortstack{RICO-Flash~\cite{wang2025rico}\\\scriptsize (FLUX.1-dev) }}
& {\shortstack{CycleCap (Ours)\\\scriptsize (SD3)}} 
& {\shortstack{CycleCap (Ours)\\\scriptsize (FLUX.1-dev)}} \\
\midrule
Obj. Category      & 73.52 & 72.61 & 70.97 &  \underline{74.17} & \textbf{77.57} \\
Obj. Number        & 60.67 & 60.00 & 57.85 &  \underline{65.12} & \textbf{65.74} \\
Obj. Color         & 89.67 & 89.33 & 88.64 & \textbf{90.36} &  \underline{89.94} \\
Spatial Relation   & 69.11 & 67.50 & 68.24 & \textbf{71.29} &  \underline{71.08} \\
Scene              & 77.07 & 76.36 & 75.56 &  \underline{78.60} & \textbf{79.06} \\
Camera Angle       & 38.00 & 39.24 & 36.90 & \textbf{40.20} &  \underline{38.42} \\
OCR                & 96.99 & 97.27 &  \underline{97.81} & \textbf{97.88} & {97.61} \\
Style              & 74.37 & 76.23 & 75.35 &  \underline{80.20} & \textbf{81.50} \\
Character Ident.   & 54.83 & 56.77 & 48.14 & \underline{58.73} & \textbf{62.67} \\
\addlinespace[3pt]
\midrule
Average            & 70.47 & 70.59 & 68.83 &  \underline{72.95} & \textbf{73.73} \\
\bottomrule
\end{tabular}\label{tab:capability_comparison}
}
\end{table}


\paragraph{\textbf{CycleCap on top of RICO-Flash}} We additionally apply CycleCap (with SD3~\cite{esser2024scaling}) on top of RICO-Flash~\cite{wang2025rico}, to examine the gains that yield from our method when deployed on an already strengthened model for captioning task through cycle consistency–ranked preference pairs. Results in Tab.~\ref{table:cyclecap_with_rico} show that CycleCap on top of RICO-Flash achieves improved performance. Note that CycleCap alone yields gains on the baseline model that surpass those of RICO-Flash. Applying CycleCap on top of RICO-Flash further improves performance, leading to the strongest performance across CompreCap and CapsBench. On CAPability, the combined approach substantially improves over RICO-Flash, which struggled to surpass the baseline, while achieving gains competitive with CycleCap alone. For MMHal, all CycleCap-based variants report similar scores, suggesting that performance on this benchmark is close to a saturation point for this training setup. Overall, the results indicate that our method is complementary to RICO-Flash approach, and that it scales effectively with captioning models boosting further their performance.

\begin{table}[t]
\centering
\caption{Performance of CycleCap applied on top of RICO-Flash~\cite{wang2025rico} across captioning and hallucination benchmarks, with Qwen2-VL-7B~\cite{wang2024qwen2} as the baseline.}
\label{table:cyclecap_with_rico}
\setlength{\tabcolsep}{4pt}
\resizebox{\columnwidth}{!}{%
\begin{tabular}{l S S S S S S S[table-format=1.2]}
\toprule
& \multicolumn{4}{c}{CompreCap} & {CAPability} & {CapsBench} & {MMHal} \\
\cmidrule(lr){2-5}
Method & {Obj} & {Attr} & {Rel} & {Uni} & & & \\
\midrule
\rowcolor{lightgray} Baseline (Qwen2-VL-7B)
& 77.86 & 2.84 & 2.80 & 61.73 
& 70.47 
& 74.17 
& 3.85 \\

RICO-Flash~
\cite{wang2025rico} 
& 79.09 & 2.93 & 2.83 & 62.93 
& 68.83 
& 75.30 
& 3.92 \\

\cyclecaprow CycleCap (Ours)
& 79.32 & 2.91 & 2.86 & 63.06 
& \textbf{73.73} 
& 76.38 
& \textbf{4.02} \\

\cyclecaprow RICO-Flash~
\cite{wang2025rico} + CycleCap (Ours) 
& \textbf{80.59} 
& \textbf{2.95} 
& \textbf{2.88} 
& \textbf{63.85} 
& {73.49 }
& \textbf{77.72} 
& 4.01 \\
\bottomrule
\end{tabular}
}
\end{table}

\subsection{Ablation Studies}

To further assess the performance and flexibility of CycleCap, we conduct ablation studies along two key factors that influence fine-tuning effectiveness: (a) the image similarity metric between the input image $x$ and its reconstruction $G(F(x))$ used for the cycle consistency reward, and (b) the text-to-image model $\mathcal{V}$ used for the backward mapping $G$.

\subsubsection{Cycle consistency metrics}
\label{ablation_metrics}

In this section, we study the effect of different image similarity metrics on CycleCap. Besides DreamSim~\cite{fu2023dreamsim}, we evaluate LPIPS~\cite{zhang2018unreasonable} and CLIP-based embeddings~\cite{radford2021learning} to measure similarity between original and reconstructed images. ollowing the setup described in Sect.~\ref{comparison_with_vanilla}, we fine-tune Qwen2-VL-2B~\cite{wang2024qwen2} for each configuration.

As shown in Tab.~\ref{table::ablation_similarity}, the DreamSim-based variant achieves the highest scores across nearly all benchmarks, indicating that perceptual metrics capturing both low- and high-level visual features provide a more effective cyclic supervision signal. LPIPS, which primarily focuses on structural similarity, yields smaller gains -- mainly on CompreCap, which emphasizes on caption completeness. CLIP similarity leads to improvements in hallucination robustness but underperforms in overall caption quality compared to DreamSim, which offers the most balanced results. These findings indicate that while CycleCap benefits from different similarity metrics, the choice of image-space metric remains a critical factor influencing its effectiveness, with more robust metrics leading to higher performance.


\begin{table}[t]
\centering
\caption{
Effect of different similarity metrics (LPIPS~\cite{zhang2018unreasonable}, CLIP~\cite{radford2021learning}, DreamSim~\cite{fu2023dreamsim}) on CycleCap performance for fine-tuning the baseline Qwen2-VL-2B~\cite{wang2024qwen2}.}
\label{table::ablation_similarity}
\setlength{\tabcolsep}{4pt}
\renewcommand{\arraystretch}{1.1}
\begin{tabular}{l 
S S S S 
S 
S[table-format=1.2]}
\toprule
& \multicolumn{4}{c}{CompreCap} & {CapsBench} & {MMHal} \\
\cmidrule(lr){2-5}
Method & {Obj} & {Attr} & {Rel} & {Uni} & & \\
\midrule
\rowcolor{lightgray} Baseline (Qwen2-VL-2B) & 73.10 & 2.72 & 2.76 & 59.35 & 69.70 & 3.63 \\
\cyclecaprow CycleCap w/ LPIPS & 74.21 & 2.68 & 2.80 & 59.70 & 64.05 & 3.49 \\
\cyclecaprow CycleCap w/ CLIP & 73.96 & 2.78 & 2.83 & 60.58 & 71.16 & \textbf{3.94}\\
\cyclecaprow CycleCap w/ DreamSim & \textbf{76.94} & \textbf{2.85} & \textbf{2.86} 
& \textbf{62.09} & \textbf{72.11} & 3.71 \\
\bottomrule
\end{tabular}
\end{table}

\subsubsection{Text-to-image models}
In this section, we study the effect of the employed text-to-image model $\mathcal{V}$ on CycleCap's performance. To this end, we replace SD3~\cite{esser2024scaling} with FLUX.1-schnell~\cite{labs2025flux1kontextflowmatching} for the reconstruction step and fine-tune Qwen2-VL-2B~\cite{wang2024qwen2} under the same training setup. The results are reported in Tab.~\ref{table::ablation_text_to_image}. Both variants yield clear improvements over the baseline model across all captioning benchmarks, confirming that the cyclic supervision scheme is robust to different reconstruction backbones. The performance of the two cases is comparable, SD3 provides slightly stronger gains on MMHal, while FLUX.1-schnell achieves slightly higher performance on CompreCap and CapsBench. Overall, these results demonstrate the flexibility of CycleCap with different text-to-image models, allowing practitioners to trade off computational efficiency, reconstruction speed, and generation style according to available resources.


\begin{table}[t]
\centering
\caption{
Effect of different text-to-image models (SD3~\cite{esser2024scaling} and FLUX.1-schnell~\cite{labs2025flux1kontextflowmatching}) on the proposed CycleCap performance for fine-tuning baseline Qwen2-VL-2B~\cite{wang2024qwen2}.}
\label{table::ablation_text_to_image}
\setlength{\tabcolsep}{4pt}
\renewcommand{\arraystretch}{1.1}
\begin{tabular}{l S S S S S S[table-format=1.2]}
\toprule
& \multicolumn{4}{c}{CompreCap} & {CapsBench} & {MMHal} \\
\cmidrule(lr){2-5}
Method & {Obj} & {Attr} & {Rel} & {Uni} & & \\
\midrule
\rowcolor{lightgray} Baseline (Qwen2-VL-2B) & 73.10 & 2.72 & 2.76 & 59.35 & 69.70 & 3.63 \\
\cyclecaprow CycleCap w/ SD3 & \bfseries 76.94 & 2.85 & 2.86 & 62.09 & 72.11 & \bfseries 3.71 \\
\cyclecaprow CycleCap w/ FLUX.1-schnell
& 75.96 & \bfseries 2.87 & \bfseries 2.89 & \bfseries 62.14 & \bfseries 72.18 & 3.65 \\
\bottomrule
\end{tabular}
\end{table}

\subsection{Qualitative Results}
\begin{figure*}[t]
    \centering
    \includegraphics[width=\textwidth]{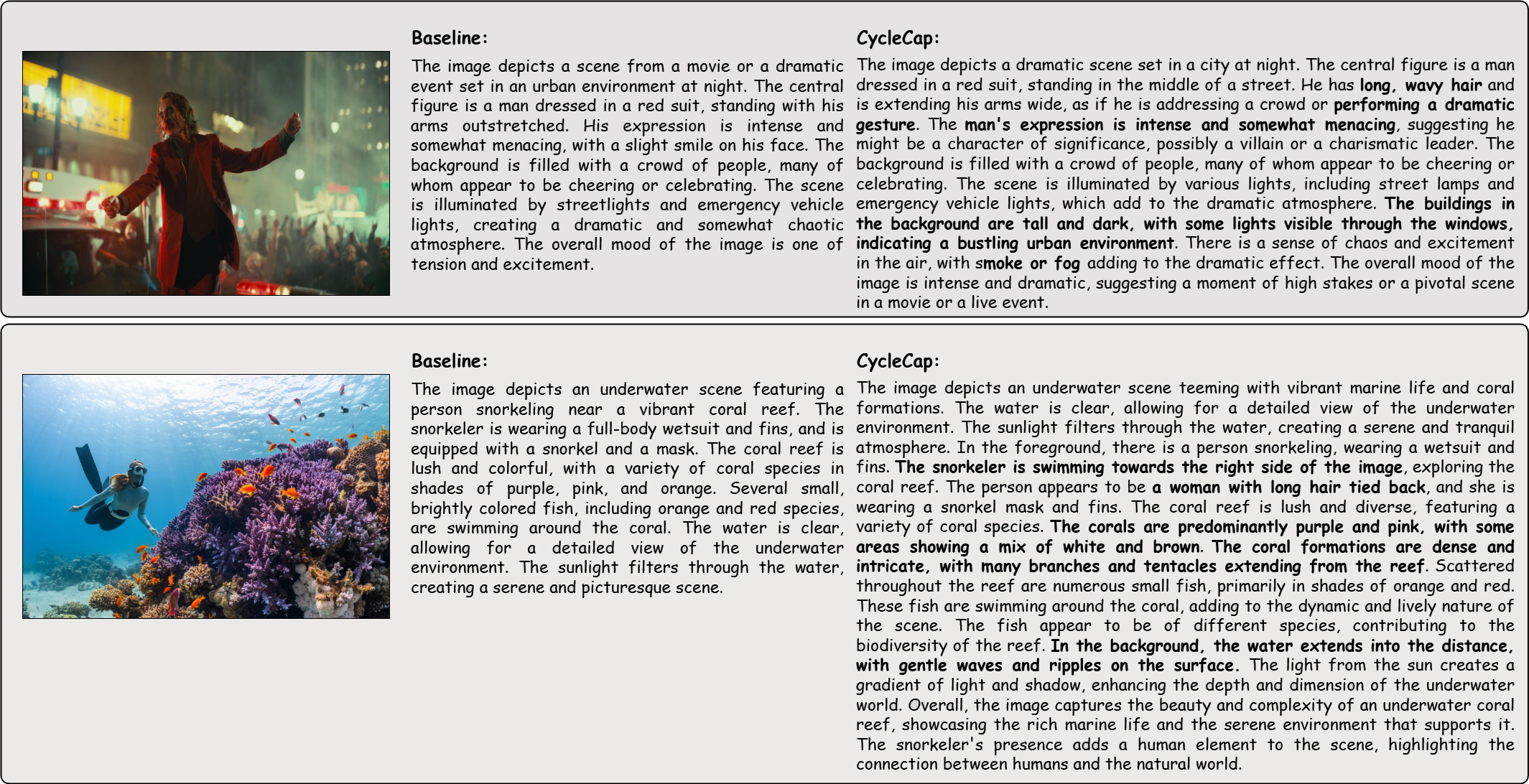}
    \caption{Qualitative comparison of captions generated by the baseline Qwen2-VL-7B~\cite{wang2024qwen2} and the proposed CycleCap on samples from the CapsBench~\cite{liu2024playground} dataset. Newly added information is indicated in \textbf{bold}. Our method produces more detailed, organized, and accurate descriptions compared to the baseline outputs.}
    \label{fig::qualitative_vanilla}
\end{figure*}

In this section, we provide qualitative results of the CycleCap method. Figure~\ref{fig::qualitative_vanilla} presents captions of sample images from the CapsBench dataset, generated with the baseline Qwen2-VL-7B~\cite{wang2024qwen2} model and the CycleCap fine-tuned version. The CycleCap captions are noticeably denser and more accurate, providing richer descriptions of spatial layout, object and character attributes, and environmental context. They also exhibit improved structure, often organizing the scene into coherent sentences covering the foreground, background, and overall atmosphere. These examples demonstrate that CycleCap improves the model’s ability to capture fine-grained visual details and produce more comprehensive captions.

\begin{figure*}[t]
    \centering
    \includegraphics[width=\textwidth]{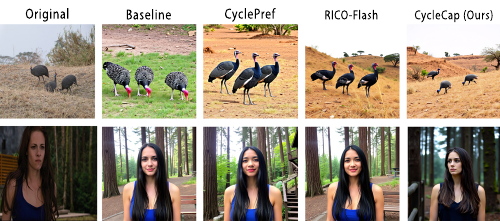}
    \caption{Qualitative comparison of image reconstructions with captions generated by SOTA and our method (CycleCap) deployed for Qwen2-VL-7B. The visualization shows that the model fine-tuned with CycleCap captures more structural details and object attributes, leading to reconstructions closer to the original image. The generated captions are provided in the Appendix.}
    \label{fig::qualitative_soa}
\end{figure*}

Additionally, Fig.~\ref{fig::qualitative_soa} presents qualitative comparisons between CycleCap (using SD3~\cite{esser2024scaling}) and the SOTA methods listed in Tab.~\ref{table::soa_comparison}. For each case, we generate an image caption and then reconstruct an image from that description. For a fair comparison, we use FLUX.1-schnell~\cite{labs2025flux1kontextflowmatching} to generate all reconstructions, since it has not been part of any of the comparative frameworks, and apply the same generation seed across all methods. The illustrations show that CycleCap's captions produce reconstructions that more closely resemble the input image in both detail and structural fidelity.

\section{Conclusion}
In this work, we introduce CycleCap, a simple yet effective fine-tuning framework for improving the captioning performance of VLMs. Our approach builds on the idea of cycle consistency, where an accurate image-to-text generation should enable a faithful text-to-image reconstruction. CycleCap uses this principle as a direct self-supervised training signal. We build a cycle consistency reward based on the similarity between the input and the reconstructed image, and combined with a GRPO-based optimization scheme it allows the model to explore multiple caption hypotheses and progressively reinforce those that yield higher image similarity. This provides a continuous, image-grounded learning signal using only raw image data, bypassing the need for costly annotated image–text datasets. Across four VLMs ranging from 1B to 7B parameters, CycleCap delivers consistent improvements on detailed captioning and hallucination benchmarks. We further show that CycleCap outperforms state-of-the-art methods that rely on cycle-consistency–derived preference datasets for fine-tuning. Unlike these approaches, which depend on the quality and scale of static preference pairs, CycleCap offers an efficient self-supervised approach that learns directly from images. Ablation studies demonstrate the flexibility of our framework -- it benefits from stronger perceptual similarity metrics and higher-fidelity generators, suggesting that the method will naturally improve further as generative models advance. We believe CycleCap opens up new opportunities for self-supervised visual–textual learning and offers a strong foundation for future research in multimodal generation. Currently, our approach is deployed for the image–text–image cycle and relies solely on image-space similarity. In future work, we will explore the extension of the training signal with cross-domain measures that, in addition to image-space similarity, assess image-text alignment across GRPO rollouts. 


\clearpage
\maketitlesupplementary
\appendix 

\renewcommand{\thefigure}{A\arabic{figure}}
\renewcommand{\thetable}{A\arabic{table}}
\renewcommand{\theequation}{A\arabic{equation}}

\setcounter{figure}{0}
\setcounter{table}{0}
\setcounter{equation}{0}


\section{Additional implementation details}
\label{sec::additional_implementation_details}
\begin{figure}[htpb]
    \centering
    \includegraphics[width=0.75\textwidth]{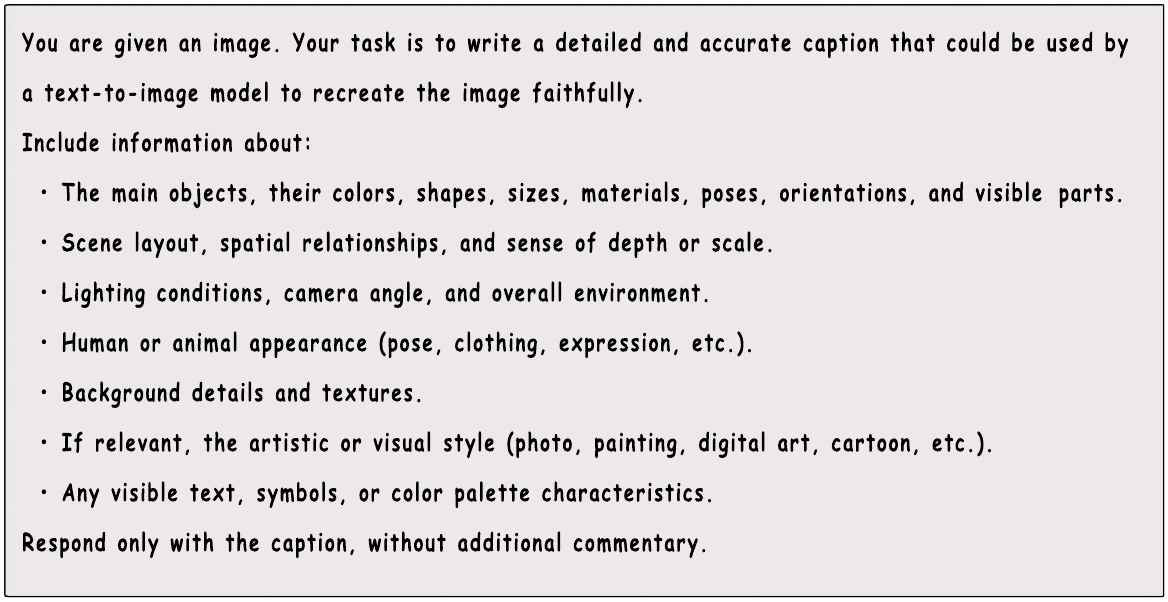}
    \caption{The designed prompt of the CycleCap framework used to generate image captions during training.}
    \label{fig::cyclecap_prompt}
\end{figure}

In Tab.~\ref{table::training_parameters} below we present the full set of training parameters used to fine-tune the models with the proposed CycleCap framework. Training time ranged from $270$ (1B) to $430$ (7B) GPU-hours on 2$\times$A100. Additionally, in Fig.~\ref{fig::cyclecap_prompt} we report the prompt used in training to generate image captions.

\begin{table}[htpb]
\centering
\caption{CycleCap training parameters.}
\label{table::training_parameters}
\setlength{\tabcolsep}{6pt}
\renewcommand{\arraystretch}{1.15}
\small
\begin{tabular}{l p{0.33\columnwidth} p{0.3\columnwidth}}
\toprule
\textbf{Category} & \multicolumn{2}{c}{\textbf{Parameters}} \\
\midrule

\multirow{2}{*}{\textbf{Training}}
& Batch size:\,64 & Learning rate:\,$10^{-5}$ \\
& Scheduler:\,Linear & Epochs:\,1 \\

\addlinespace
\multirow{2}{*}{\textbf{LoRA}}
& Rank $r$:\,64 & Dropout:\,0.05 \\
& Target modules:\,all linear projection layers & Vision tower:\,frozen \\

\addlinespace
\multirow{2}{*}{\textbf{GRPO}}
& KL weight $\beta$:\,0.04 & Clip value $\varepsilon$:\,0.02 \\
& Generations $n$:\,8 & \\

\addlinespace
\multirow{2}{*}{\textbf{Optimization}}
& Optimizer:\,AdamW & Precision:\,bfloat16 \\
& Gradient checkpointing & \\

\bottomrule
\end{tabular}
\end{table}

\section{Additional experiments}
\subsection{Evaluation on visual understanding and reasoning benchmarks}

Our main objective is to use the CycleCap fine-tuning framework to improve captioning performance. Hence, enhancing performance on broader visual-language understanding, reasoning, or complex VQA tasks lies outside the scope of this work. Nevertheless, for the sake of completeness, we examine how CycleCap affects these abilities by evaluating the models presented in Tab.~1 in the main paper, on a set of widely used benchmarks that include a variety of visual inputs, ranging from natural images to maps and charts, and that measure visual-language capabilities in high-level perception and reasoning tasks. Tab.~\ref{table::comparison_vanilla_understanding_reasoning} summarizes the results. We observe that the CycleCap fine-tuned models exhibit comparable performance to their baseline counterparts. This indicates that CycleCap does not harm the generic multimodal capabilities of the models, and can even lead to small improvements, despite being trained solely on the captioning task.

\begin{table}[htpb]
\centering
\caption{Comparison of CycleCap with baseline models on visual-language understanding and reasoning benchmarks. The evaluated models are the same as those reported in Tab.~\ref{table::comparison_vanilla_captioning} in the main paper.}
\label{table::comparison_vanilla_understanding_reasoning}
\setlength{\tabcolsep}{5pt}
\renewcommand{\arraystretch}{1.15}
\footnotesize
\begin{adjustbox}{max width=\textwidth}
\begin{tabular}{
l
S[table-format=4.2]
S[table-format=2.2]
S[table-format=2.2]
S[table-format=2.2]
S[table-format=2.2]
}
\toprule
{Model}
& {MME$_{\mathrm{sum}}$~\cite{fu2023mme}}
& {MMBench$_{\mathrm{test}}$~\cite{liu2024mmbench}}
& {MMStar~\cite{chen2024we}}
& {MMMU$_{\mathrm{val}}$~\cite{yue2024mmmu}}
& {Hall-Bench~\cite{guan2024hallusionbench}} \\
\midrule

\multicolumn{6}{c}{\textit{InternVL3-1B}~\cite{zhu2025internvl3}} \\ \hline
Baseline & 1873.19 & 70.51 & \bfseries 52.31 & \bfseries 40.22 & 47.21 \\
\cyclecaprow CycleCap (Ours) & \bfseries 1892.90 & \bfseries 71.41 & 52.00 & 39.33 & \bfseries 49.31 \\
\addlinespace[3pt]
\hline

\multicolumn{6}{c}{\textit{Qwen2-VL-2B}~\cite{wang2024qwen2}} \\ \hline
Baseline & 1866.25 & 71.08 & 43.34 & 40.44 & 50.68 \\
\cyclecaprow CycleCap (Ours) & \bfseries 1872.50 & \bfseries 71.97 & \bfseries 43.70 & \bfseries 40.89 & \bfseries 51.10 \\
\addlinespace[3pt]
\hline

\multicolumn{6}{c}{\textit{Qwen2.5-VL-3B}~\cite{bai2025qwen2_5}} \\ \hline
Baseline & 2147.22 & 77.57 & \bfseries 56.01 & 45.89 & \bfseries 57.51 \\
\cyclecaprow CycleCap (Ours) & \bfseries 2148.82 & \bfseries 78.75 & 55.04 & \bfseries 46.22 & 56.04 \\
\addlinespace[3pt]
\hline

\multicolumn{6}{c}{\textit{Qwen2-VL-7B}~\cite{wang2024qwen2}} \\ \hline
Baseline & \bfseries 2294.35 & 78.30 & 57.28 & 50.44 & 57.93 \\
\cyclecaprow CycleCap (Ours) & 2293.37 & \bfseries 78.81 & \bfseries 58.18 & \bfseries 51.11 & \bfseries 58.88 \\

\bottomrule
\end{tabular}
\end{adjustbox}
\end{table}

\subsection{Analysis of Caption Length}

In Tab.~\ref{table::avg_token_length} we report the average length (in tokens) of the captions generated by each model shown in Tab.~\ref{table::comparison_vanilla_captioning} in the main paper, on the Comprecap~\cite{lu2025benchmarking} benchmark. The results show that CycleCap fine-tuned models produce caption lengths comparable to baseline models for all cases. This suggests that the improvements in captioning performance are not solely associated with increased verbosity, but rather reflect qualitative improvements in the generated descriptions, such as better structure and stronger semantic alignment with the visual content.

\begin{table}[htpb]
\centering
\caption{Average caption length (in tokens) on the CompreCap~\cite{lu2025benchmarking} benchmark.}
\label{table::avg_token_length}
\setlength{\tabcolsep}{6pt}
\resizebox{\columnwidth}{!}{
\begin{tabular}{l c c c c c c c c}
\toprule
& \multicolumn{2}{c}{InternVL3-1B~\cite{zhu2025internvl3}} 
& \multicolumn{2}{c}{Qwen2-VL-2B~\cite{wang2024qwen2}} 
& \multicolumn{2}{c}{Qwen2.5-VL-3B~\cite{bai2025qwen2_5}} 
& \multicolumn{2}{c}{Qwen2-VL-7B~\cite{wang2024qwen2}} \\
\cmidrule(lr){2-3} \cmidrule(lr){4-5} \cmidrule(lr){6-7} \cmidrule(lr){8-9}
 & Baseline & CycleCap & Baseline & CycleCap & Baseline & CycleCap & Baseline & CycleCap \\
Avg. Length
& 152.64 & 179.80 
& 209.39 & 202.73 
& 176.06 & 198.73 
& 384.56 & 385.23 \\
\bottomrule
\end{tabular}
}
\end{table}

\subsection{Effect of GRPO's number of generations}
We analyze the sensitivity of CycleCap to the number of generations, $n$, in GRPO~\cite{shao2024deepseekmath}, which corresponds to the number of generated captions per image during training to compute the relative advantage in (\ref{eq::advantage}). To this end, we fine-tune the Qwen2-VL-7B~\cite{wang2024qwen2} model with CycleCap for $n = 2, 4$ (in addition to $n = 8$) using SD3~\cite{esser2024scaling}, following a similar process to the one described in Section 4.1 of the main paper. Tab.~\ref{table::group_size_ablation} reports CycleCap's performance for different values of $n$ on captioning benchmarks.

\begin{figure}[htpb]
    \centering
    \includegraphics[width=0.9\textwidth]{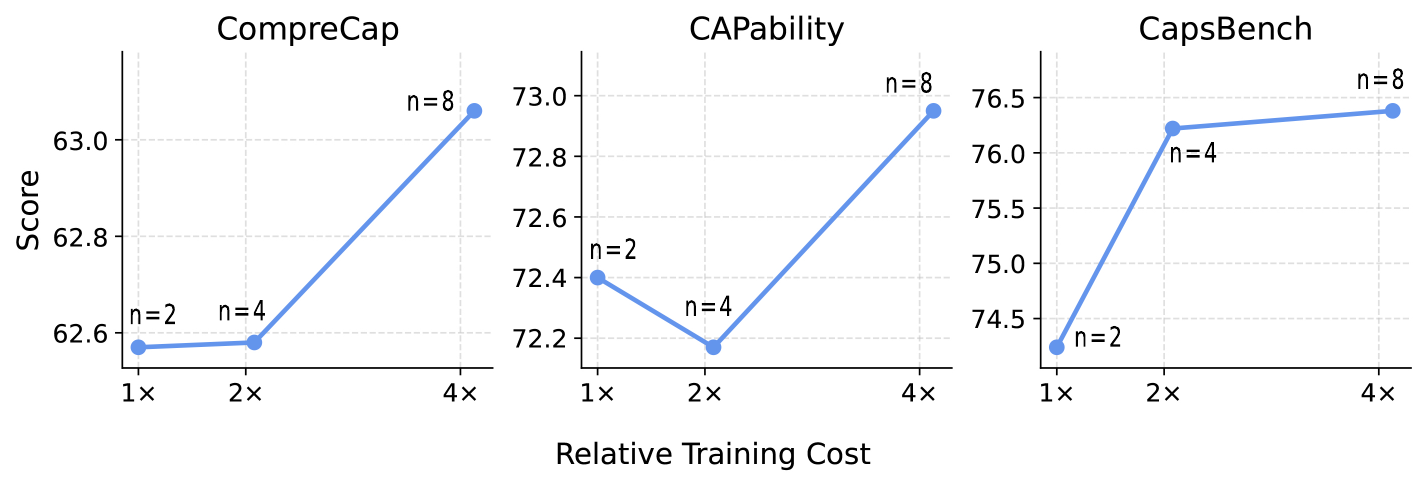}
    \caption{Effect of the number of GRPO generations $n$ on CycleCap performance vs. relative training cost. We report performance on CompreCap (Unified Score)~\cite{lu2025benchmarking}, CAPability~\cite{liu2025capability}, and CapsBench~\cite{liu2024playground} benchmarks. Relative training cost is normalized with respect to $n=2$ ($1\times$).}
    \label{fig::group_vs_cost}
\end{figure}

Across all configurations, CycleCap consistently improves over the baseline. Notably, even with $n = 2$ or $n = 4$, CycleCap surpasses SOTA approaches in most cases (see Tab.~\ref{table::soa_comparison} in the main paper), emphasizing the gains that arise from the GRPO-based training objective. Increasing the number of caption generations to $n = 8$ achieves the best overall results across the evaluated benchmarks. In Fig.~\ref{fig::group_vs_cost}, we illustrate the relationship between CycleCap performance and training cost for different values of $n$. As shown, increasing $n$ scales approximately linearly with training cost, with larger numbers of generated captions leading to improved performance.

\begin{table}[htpb]
\centering
\caption{Effect of the number of GRPO generations $n$ (captions generated per image) on CycleCap performance on captioning benchmarks CompreCap (Unified Score)~\cite{lu2025benchmarking}, CAPability~\cite{liu2025capability}, and CapsBench~\cite{liu2024playground}.}
\label{table::group_size_ablation}
\setlength{\tabcolsep}{8pt}
\resizebox{0.9\textwidth}{!}{
\begin{tabular}{l c c c c}
\toprule
Method & Generations $n$ & CompreCap & CAPability & CapsBench \\
\midrule
\rowcolor{lightgray} Baseline (Qwen2-VL-7B) & --- & 61.73 & 70.47 & 74.17 \\
CycleCap & 2 & 62.57 & 72.40 & 74.24 \\
CycleCap & 4 & 62.58 & 72.17 & 76.22 \\
CycleCap & 8 & \textbf{63.06} & \textbf{72.95} & \textbf{76.38} \\
\bottomrule
\end{tabular}
}
\end{table}

\subsection{Controlled comparison with DPO on CyclePrefDB-I2T}

In Section 4.4 in the main paper, we compare CycleCap with prior cycle-consistency-based methods, including CyclePref~\cite{bahng2025cycle}, which applies DPO~\cite{rafailov2023direct} using the CyclePrefDB-I2T~\cite{bahng2025cycle} preference dataset. Since the models reported in Tab.~\ref{table::soa_comparison} in the main paper are trained on different datasets, we additionally evaluate both approaches using the same training data for a more controlled comparison.


Specifically, we fine-tune Qwen2-VL-7B~\cite{wang2024qwen2} with CycleCap using the images from CyclePrefDB-I2T (approximately 7k images). In CycleCap training, a group of $n=8$ candidate captions is generated for each image during GRPO optimization. Since CyclePrefDB-I2T~\cite{bahng2025cycle} originally contains about 55 preference pairs per image, we randomly sample 8 preference pairs per image and use this subset for DPO fine-tuning, so that both methods receive a comparable number of supervision signals per image. We fine-tune each model for one epoch using identical optimization settings (Tab.~\ref{table::training_parameters}) and report the results in Tab.~\ref{table::comparison_on_cyclepref}.

DPO fine-tuning with the sub-sampled CyclePrefDB-I2T~\cite{bahng2025cycle} pairs leads to modest improvements over the baseline model. CycleCap, however, consistently achieves stronger gains across the adopted evaluation benchmarks. More specifically, by generating candidate captions during training and optimizing them through cycle consistency rewards, the proposed method provides a stronger learning signal than relying solely on static preference pairs from the dataset.

\begin{table*}[htpb]
\centering
\caption{Controlled comparison between DPO and CycleCap using the sub-sampled CyclePrefDB-I2T~\cite{bahng2025cycle} training set with Qwen2-VL-7B~\cite{wang2024qwen2} as the base model. Results are reported on CompreCap (Unified Score)~\cite{lu2025benchmarking}, CAPability~\cite{liu2025capability}, CapsBench~\cite{liu2024playground}, and MMHal~\cite{sun2024aligning} benchmarks.}
\label{table::comparison_on_cyclepref}
\setlength{\tabcolsep}{8pt}
\resizebox{0.9\textwidth}{!}{
\begin{tabular}{l c c c c}
\toprule
Method & CompreCap & CAPability & CapsBench & MMHal \\
\midrule
\rowcolor{lightgray}
Baseline (Qwen2-VL-7B) & 61.73 & 70.47 & 74.17 & 3.85 \\
DPO & 61.78 & 71.11 & 74.41 & 3.95 \\
\cyclecaprow CycleCap (Ours) & \textbf{62.33} & \textbf{71.89} & \textbf{74.47} & \textbf{4.01} \\
\bottomrule
\end{tabular}
}
\end{table*}

\subsection{Detailed results of Table~\ref{table::soa_comparison}}
Tab.~\ref{table::soa_comparison_detailed} provides the detailed evaluation results corresponding to Tab.~\ref{table::soa_comparison} in the main paper.

\begin{table*}[htpb]
\centering
\caption{Detailed results corresponding to Tab.~\ref{table::soa_comparison} in the main paper, comparing the proposed CycleCap with SOTA cycle-consistency-based approaches.}
\label{table::soa_comparison_detailed}
\setlength{\tabcolsep}{6pt}
\resizebox{\textwidth}{!}{%
\begin{tabular}{l c c c c c c c c}
\toprule
& & \multicolumn{4}{c}{CompreCap} & {CAPability} & {CapsBench} & {MMHal} \\
\cmidrule(lr){3-6}
Method & T2I Model & {Obj} & {Attr} & {Rel} & {Uni} & & & \\
\midrule
\rowcolor{lightgray}
Baseline (Qwen2-VL-7B)
& --- 
& 77.86 & 2.84 & 2.80 & 61.73
& 70.47
& 74.17 & 3.85 \\

CyclePref~\cite{bahng2025cycle}
& SD3 
& 77.29 & 2.85 & 2.85 & 62.03
& 70.59
& 74.27 & 3.95 \\

RICO-Flash~\cite{wang2025rico}
& FLUX.1-dev 
& 79.09 & 2.93 & 2.83 & 62.93
& 68.83
& 75.30 & 3.92 \\

\midrule
\cyclecaprow CycleCap (Ours)
& SD3 
& 79.32 & 2.91 & 2.86 & 63.06
& 72.95
& 76.38 & 4.02 \\

\cyclecaprow CycleCap (Ours)
& FLUX.1-dev 
& \textbf{79.67} & \textbf{2.93} & \textbf{2.90} & \textbf{63.64}
& \textbf{73.73}
& \textbf{77.25} & \textbf{4.02} \\
\bottomrule
\end{tabular}
}
\end{table*}

\section{Additional qualitative comparisons}
\begin{figure*}[htpb]
    \centering
    \includegraphics[width=\textwidth]{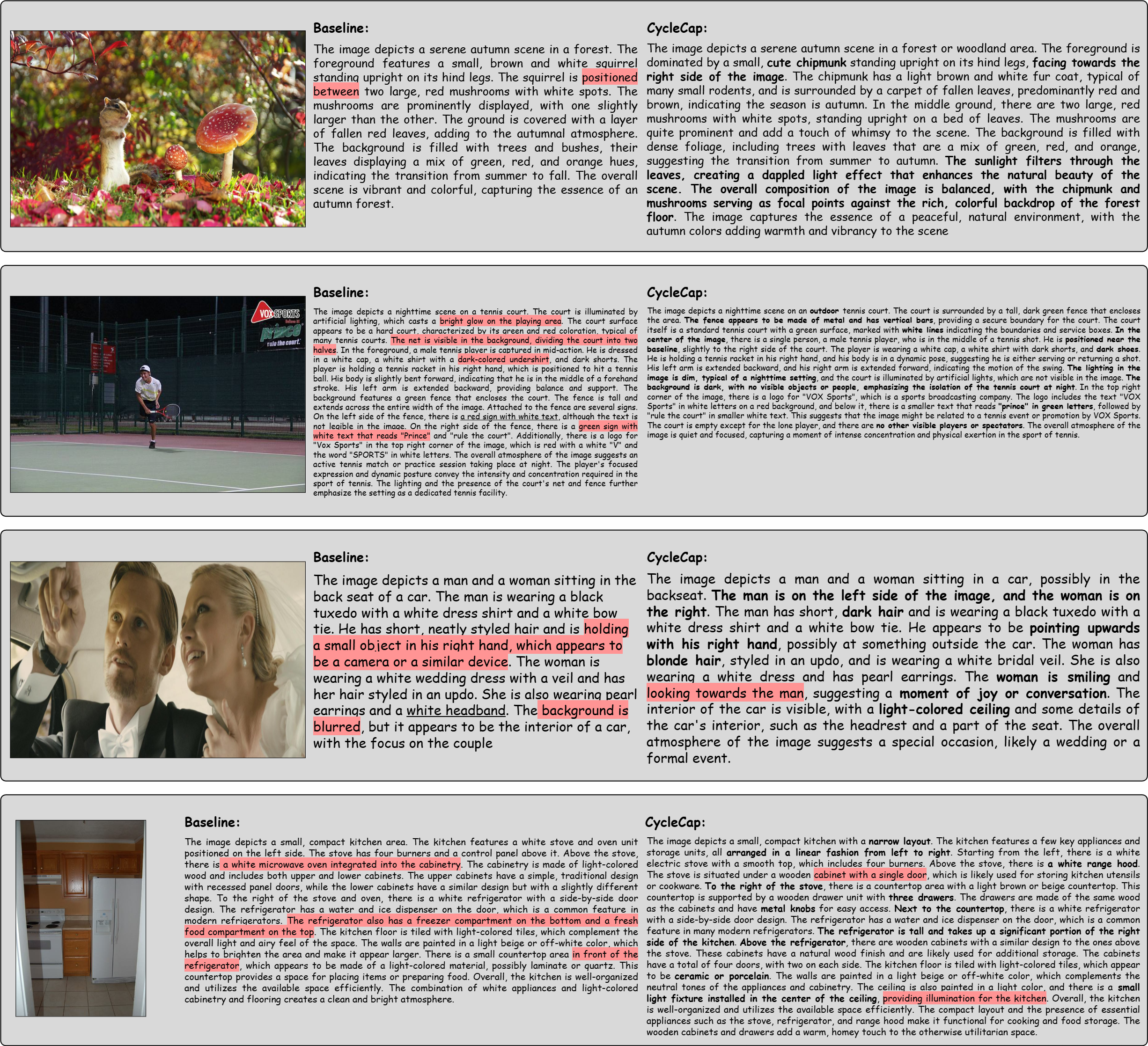}
    \caption{Additional qualitative comparisons between captions generated by the baseline Qwen2-VL-7B~\cite{wang2024qwen2} model and the CycleCap fine-tuned version, for image samples from the CapsBench~\cite{liu2024playground} and CompreCap~\cite{lu2025benchmarking} benchmarks. Newly added information is shown in \textbf{bold}, content present in the baseline but not preserved in the CycleCap caption is \underline{underlined}, and hallucinations are highlighted in pink. Overall, CycleCap tends to produce more structured and grounded descriptions, while it adds details related to object attributes, spatial layout, and broader scene context.}
    \label{fig::qualitative_extra}
\end{figure*}

\begin{figure*}[htpb]
    \centering
    \includegraphics[width=0.75\textwidth]{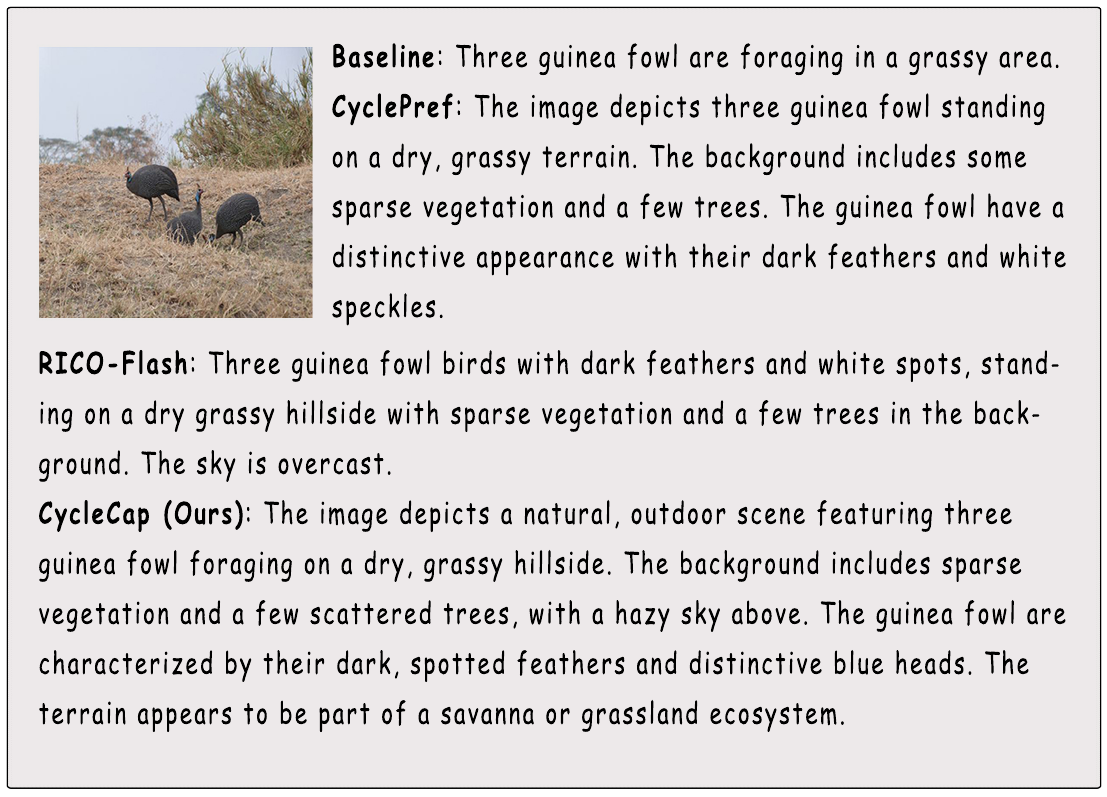}
    \caption{Generated captions corresponding to the qualitative reconstruction comparison shown in the first row of Fig.~\ref{fig::qualitative_soa} in the main paper.}
    \label{fig::qualitative_soa_captions_1}
\end{figure*}

\begin{figure*}[htpb]
    \centering
    \includegraphics[width=0.75\textwidth]{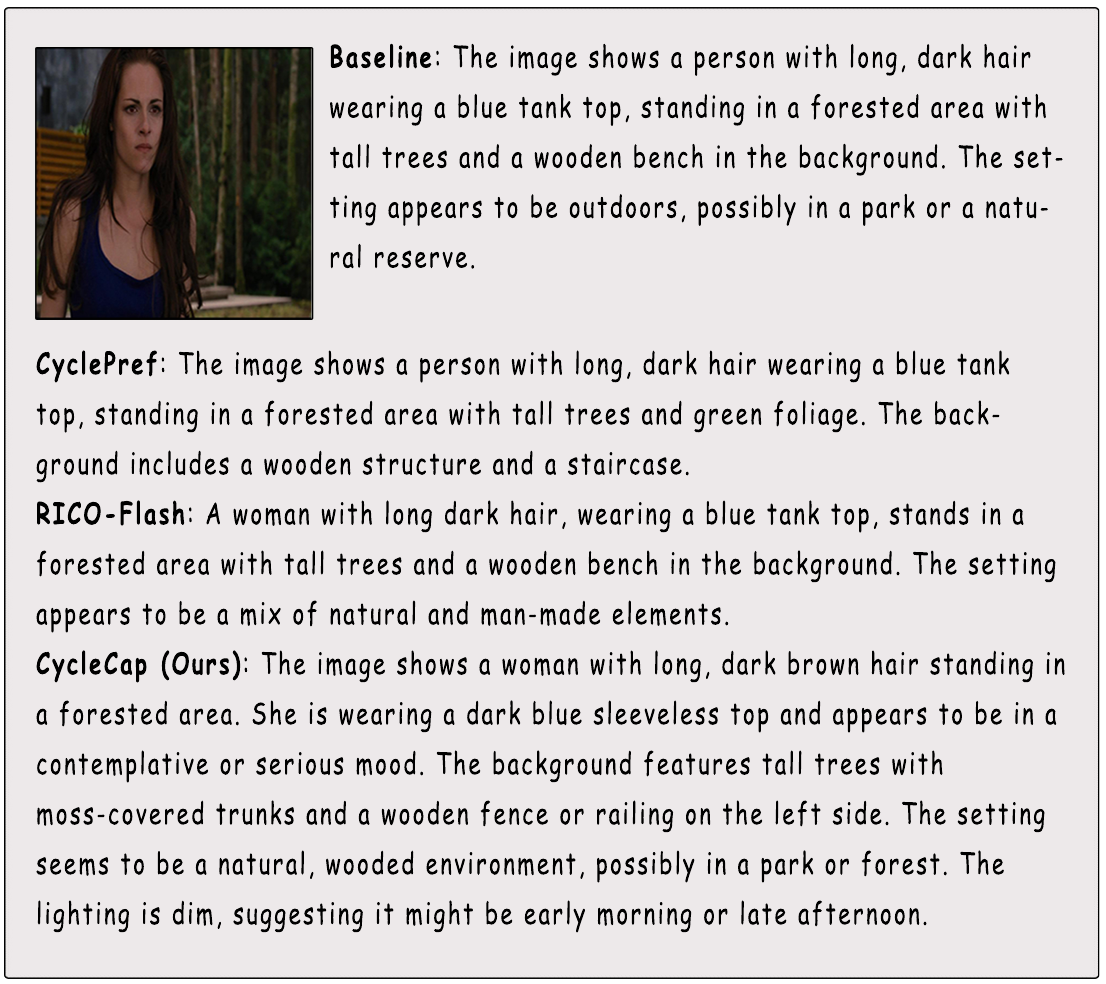}
    \caption{Generated captions corresponding to the qualitative reconstruction comparison shown in the second row of Fig.~\ref{fig::qualitative_soa} in the main paper.}
    \label{fig::qualitative_soa_captions_2}
\end{figure*}

In Fig.~\ref{fig::qualitative_extra} we present additional qualitative comparisons between captions generated by the baseline Qwen2-VL-7B~\cite{wang2024qwen2} and its CycleCap fine-tuned variant. Moreover, in Fig.~\ref{fig::qualitative_soa_captions_1} and \ref{fig::qualitative_soa_captions_2} we present the generated captions of each method that correspond to the qualitative comparison of Fig.~\ref{fig::qualitative_soa} in the main paper. As shown, CycleCap produces more detailed and accurate captions compared to SOTA methods.

%
%
\clearpage
\bibliographystyle{splncs04}
\bibliography{main}

\end{document}